\documentclass[conference]{IEEEtran}
\IEEEoverridecommandlockouts
\usepackage{cite}
\usepackage{amsmath,amssymb,amsfonts}
\usepackage{algorithmic}
\usepackage{graphicx}
\usepackage{textcomp}
\usepackage{xcolor}
\usepackage{comment}
\usepackage{multirow}
\usepackage{makecell}
\usepackage{marvosym}
\def\BibTeX{{\rm B\kern-.05em{\sc i\kern-.025em b}\kern-.08em
    T\kern-.1667em\lower.7ex\hbox{E}\kern-.125emX}}
\begin{document}

\title{A New Approach to Training Multiple Cooperative Agents for Autonomous Driving}

\author{\IEEEauthorblockN{Ruiyang Yang, Siheng Li, Beihong Jin\textsuperscript{\Letter}}
\IEEEauthorblockA{\textit{State Key Laboratory of Computer Science, Institute of Software, Chinese Academy of Sciences, Beijing 100190, China} \\
\textit{University of Chinese Academy of Sciences, Beijing 100190, China} \\
Email: \textsuperscript{\Letter}beihong@iscas.ac.cn}
}

\maketitle

\begin{abstract}
Training multiple agents to perform safe and cooperative control in the complex scenarios of autonomous driving has been a challenge. For a small fleet of cars moving together, this paper proposes Lepus, a new approach to training multiple agents. Lepus adopts a pure cooperative manner for training multiple agents, featured with the shared parameters of policy networks and the shared reward function of multiple agents. In particular, Lepus pre-trains the policy networks via an adversarial process, improving its collaborative decision-making capability and further the stability of car driving. Moreover, for alleviating the problem of sparse rewards, Lepus learns an approximate reward function from expert trajectories by combining a random network and a distillation network. We conduct extensive experiments on the MADRaS simulation platform. The experimental results show that multiple agents trained by Lepus can avoid collisions as many as possible while driving simultaneously and outperform the other four methods, that is, DDPG-FDE, PSDDPG, MADDPG, and MAGAIL(DDPG) in terms of stability.
\end{abstract}

\begin{IEEEkeywords}
autonomous driving, reinforcement learning, imitation learning
\end{IEEEkeywords}

\section{Introduction}
The past few years have witnessed a surge of development in autonomous driving technology. However, autonomous driving in complex scenarios still has a long way to go before the technology becomes mature and safe enough. Typically, autonomous driving of multiple cars has been a challenge, where each car only has access to the local information it observes, but not to the information and actions of other cars. These cars need to cooperate so as to drive safely. Reinforcement learning methods, which can optimize sequential decisions, have shown excellent performance on the driving decision-making of a single car agent~\cite{2017Deep}~\cite{2018End}. Imitation learning methods, which directly learn the behavior of a driver from recorded driving experiences and generate the driving policies, provide alternative solutions to autonomous driving ~\cite{li2017infogail, lee2020mixgail, fei2020triple}. Therefore, multi-agent systems are expected to be constructed to achieve autonomous driving through multi-agent reinforcement learning~\cite{grigorescu2020survey} or imitation learning~\cite{hussein2017imitation}.

The cooperative mode and training goal of the agents are the basis of autonomous driving oriented multi-agent system. Nowadays, there are three cooperative modes for multiple agents~\cite{zhang2021multi}. (1) Fully decentralized mode. In this mode, a reinforcement learning algorithm for training a single agent is directly applied to the multi-agent scenarios, in which each agent treats all other agents as a part of the environment and is still trained individually with the goal of maximizing its own cumulative reward. DDPG-FDE~\cite{lillicrap2015continuous} adopts this mode. However, agents trained individually without communicating with other agents clearly cannot perform effectively in the multi-agent scenarios. (2) Fully centralized mode. In this mode, each agent sends its own state to the central controller. The central controller makes the action decisions for each agent based on all the information it receives, and then transmits the decisions to each agent for execution. In this way, the central controller is expected to maximize the obtained cumulative rewards. PS-DDPG~\cite{kaushik2019parameter} adopts this mode. However, the joint action space of multiple agents grows exponentially as the number of agent increases. This mode is also not effective in the multi-agent scenarios. (3) Centralized training and decentralized execution mode. In this mode, the central controller only works during training. When the training is done, each agent makes action decisions itself based on the policy network trained using the central controller. Nash equilibrium~\cite{gibbons1992primer} is introduced in this mode as the training goal for multi-agent reinforcement learning, i.e., training will end if each agent cannot obtain a higher reward by adjusting its policy when other agents in the environment do not change their policies. In this way, one or some agents may obtain the optimal local cumulative reward, i.e., not necessarily every agent can obtain the optimal local cumulative reward. MADDPG (Multi-Agent Deep Deterministic Policy Gradient)~\cite{lowe2017multi} adopts this mode.

We note that there is a specific kind of scenarios in autonomous driving, in which multiple cars drive together to reach the same destination probably for logistics transportation or for group travelling or for other purposes. Existing methods do not examine the distinctiveness of these scenarios and do not perform well in such scenarios due to high training cost and low driving stability.

We hold the view that in the multi-car scenarios described above (i.e., the cooperative multi-car scenarios), there is a strong cooperative relationship among the cars, which should be utilized to propose an effective multi-agent training method. In the paper, we propose Lepus. In Lepus, we follow the centralized training and decentralized execution mode for multi-car agents and adopt a pure cooperative manner for the multi-agent training. Meanwhile, we adopt inverse reinforcement learning, an imitation learning technique, to learn reward functions from expert trajectory data to alleviate the sparse reward problem.

The contributions of this paper are summarized as follows:

\begin{itemize}
\item For the cooperative multi-car scenarios, we adopt a pure cooperative manner for multi-agent training where policy networks share parameters and agents share a reward function. In particular, policy networks equipped with shared parameters are pre-trained via an adversarial process, with the aim of reducing the computational cost for training.
\item To fit with the pure cooperative manner, we design the corresponding distillation network and the random network, and make them cooperate with each other to learn the shared approximate reward function from joint expert trajectories. The resulting rewards, further adjusted by driving rules, help to make the training progress of the policy networks smoothly.
\item With the aid of the autonomous driving simulation platform MADRaS, we compare the performance of Lepus with PS-DDPG, MADDPG, DDPG-FDE and MAGAIL(DDPG). The experimental results show that Lepus outperforms the other four methods in terms of stability.
\end{itemize}

The remainder of this paper is organized as follows. The related work is summarized in Section $II$. Section $III$ gives the problem formulation. Section $IV$ describes Lepus in detail. Section $V$ evaluates Lepus through extensive experiments. Finally, we conclude this paper in Section $VI$.

\section{Related Work}
Our work is related to two topics: multi-agent reinforcement learning and multi-agent imitation learning.

In multi-agent reinforcement learning, some methods adopt the fully decentralized mode. Taking IQL (Independent Q-learning)~\cite{tan1993multi} as an example, each agent in IQL runs the Q-learning algorithm independently. Some methods adopt the fully centralized mode. For example, the PS-DDPG method contains a policy network with shared network parameters to generate policies for multiple agents. Some methods adopt the centralized training and decentralized execution mode. For example, the COMA (Counterfactual Multi-Agent) method~\cite{foerster2018counterfactual} adopts the actor-critic mode for multiple agents, that is, an agent needs the state-action pairs from all other agents as input when employing the critic to evaluate Q values, and each agent trains and updates a corresponding actor using the local observation of its own as input. In the COMA method, all agents cooperate with each other to accomplish a task with a shared reward function. Moreover, COMA also evaluate the contributions of each agent to the overall task based on the counterfactual thinking. The MADDPG method is a multi-agent version of the DDPG algorithm. Instead of simply combining multiple agents trained by the DDPG algorithm, MADDPG adopts the centralized training and decentralized execution mode, allowing the critic network of each agent to be trained based on the current state and action information of other agents. Further, R-MADDPG (Recurrent MADDPG)~\cite{wang2020r} is an extension of MADDPG. It adds a recurrent neural network structure (i.e., LSTM) to MADDPG in three different ways. The first one is to introduce LSTM to the network structure of the actor. The second one is to introduce LSTM to the network structure of the critic. And the last one is to introduce LSTM to the network structures of both the actor and the critic. The purpose is to coordinate multiple agents for cooperative training when the communication is limited.

In multi-agent imitation learning, multi-agent inverse reinforcement learning~\cite{natarajan2010multi} supposes that there is a centralized mediator to coordinate the actions of each agent in the scenario to obtain the optimal policies for the whole scenario. That is, each agent learns its own reward function using expert trajectory data. After forming the joint reward, the multi-agent overall policies are optimized under the control of the centralized mediator. Aiming at multiple Nash equilibrated and non-stationary environment in the multi-agent system, MAGAIL inherits the idea of GAN~\cite{goodfellow2014generative}, in which the generator controls the policies of all the agents in a distributed manner, while the discriminator is trained to distinguish the differences between the actions produced by each agent and its corresponding expert actions. MAGAIL relies on the feature of the application (cooperation or competition) as prior knowledge to change the structure of the discriminator. In the multi-agent scenarios, the agents keep interacting with each other. Therefore, the environmental states as the input in the training and testing phases might be  sampled from different distributions, thus leading to the covariate shift problem, i.e., during testing phase, the performance differences between the trained agent and the real data become progressively larger. PS-GAIL~\cite{bhattacharyya2018multi} combines GAIL~\cite{ho2016generative} and PS-TRPO (Parameter Sharing TRPO) to solve the problem described above, making it possible to learn policies for controlling multiple car agents simultaneously in the multi-agent autonomous driving scenarios.

Compared with existing multi-agent training methods for the autonomous driving, Lepus adopts a pure cooperative manner among agents for the cooperative multi-car scenarios, and then pre-trains policy networks using an adversarial process. Lepus provides a way to learn approximate reward function from joint expert trajectories, which reduces the network parameters, and improves the driving stability as well as the decision-making ability.

\section{Formulation}
For the cooperative multi-car scenarios, we propose that cars run in a pure cooperative manner and our task is to train multiple car agents so as to drive collaboratively while avoiding collisions as many as possible.

Specifically, we suppose that there are $M$ car agents to be trained in the environment. Given a time step $t$, let $s_t^i$ be the state of the $i$-th car agent at time step $t$, and $a_t^i$ be the action taken by the $i$-th car agent at time step $t$. Next, $S=\left\{s_t^i\right\}, i\in\left\{1,2,...,M\right\}$ denotes the joint state of multiple agents at time step $t$, and $A=\left\{a_t^i\right\}, i\in\left\{1,2,...,M\right\}$ denotes the joint action of multiple agents at time step $t$. Then, $\tau_E=\left\{(s_t,a_t )^1,...,(s_t,a_t )^M \right\}_{t=0,1,2,...}$ denotes the  joint trajectory of multiple driving experts.

Given $\tau_E$, we will learn the approximate reward function, denoted by $R$. In particular, in the pure cooperative manner, the approximate reward function $R$ will be shared by multiple car agents.

Our goal of autonomous driving is to train the multiple car agents to find the joint optimal policy $Joint_\mu=\left\{(\mu^i)^*\right\}$, where the optimal policy $(\mu^i)^*$ of the car agent $i$ is a mapping from the observed local state $s_t^i$ to the action $a_t^i$ that it takes, denoted by $a_t^i=\mu(s_t^i)$. Since cars are in the pure cooperative manner, the process of finding the joint optimal policy is actually to maximize the cumulative expected reward $U$ of multiple car agents, denoted by \eqref{eq1}
described below.

\begin{equation}
\begin{split}
\begin{aligned}
max\left ( U \right ) =\sum_{i=1}^MU_i
\label{eq1}
\end{aligned}
\end{split}
\end{equation}

\noindent where $U_i=\sum_{t=0}^T\gamma^t R(s_t^i,\mu(s_t^i))$, $\gamma\in[0,1]$ is the discount factor.

Given that the simulation platform MADRaS (Multi-Agent Driving Simulator)~\cite{santara2020madras} is used to simulate autonomous driving scenarios, the input state $s$ is a $65$-dimensional vector obtained from the sensors on each car agent, and the action $a$ is composed of a $3$-dimensional continuous vector [steering, acceleration, braking].

\section{Lepus Approach}
\begin{figure*}[htbp]
\centering{\includegraphics[width=0.8\textwidth]{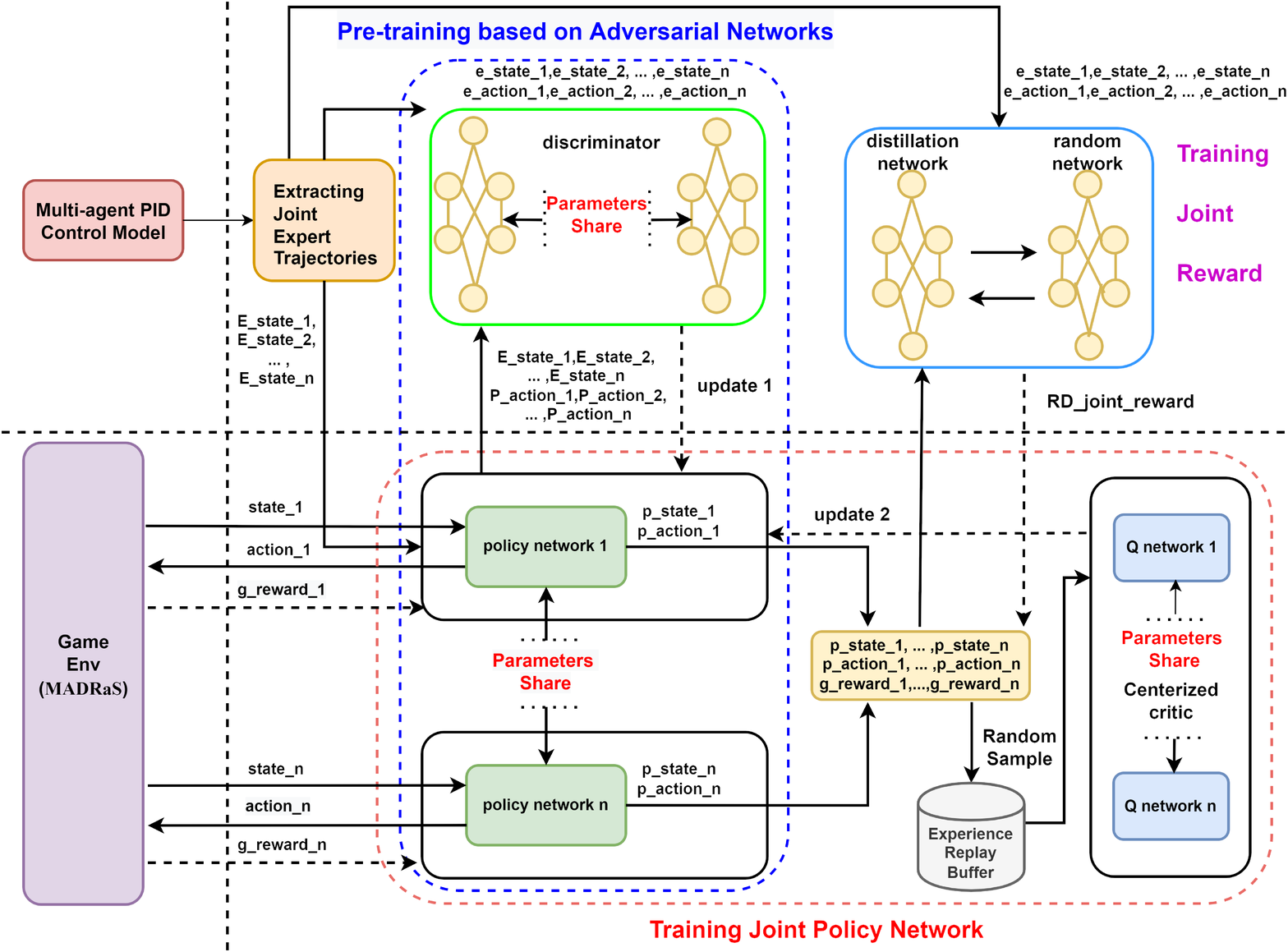}}
\caption{Framework of Lepus.}
\label{fig1}
\end{figure*}

The Lepus approach, as shown in Fig.~\ref{fig1}, consists of three parts: learning a joint reward function from joint expert trajectories, pre-training the policy networks using an adversarial process and training the shared parameters of policy networks. After training, policy networks with shared parameters will act as car agents to make rational decisions for a given state.

\subsection{Learning a Joint Reward Function for Multiple Car Agents}
To fit with the pure cooperative driving mode of agents, we first construct a joint random distillation network to provide a shared reward function (i.e., $RD\_joint\_reward$ in Fig.~\ref{fig1}) for the cooperative driving of the multiple car agents. 

The joint random distillation network consists of two parts: a distillation network $f_{\hat{\theta}}\left(S, A|\tau_{E}\right)$ and a random network $f_{\theta}\left(S, A|\tau_{E}\right)$. Specifically, the distillation network contains an input layer, four fully connected layers and an output layer. The numbers of hidden units of the four fully connected layers are all 128, and LeakyReLU is used as the activation function. The random network contains an input layer, an output layer and a fully connected layer with 128 hidden units and LeakyReLU as the activation function. 

The specific training process of the random distillation network is as follows. First, we randomly set parameter values of the random and distillation networks. Then, keeping the parameters of the random network fixed, we feed the same joint expert trajectory data (i.e., $\left.\left(S, A|\tau_{E}\right)\right)$ into both the distillation network and the random network and update the parameters of the distillation network continuously in given iterations, so that the output of the distillation network approaches the output of the random network gradually. 

When the training is done, the whole joint random distillation network is used to constitute a reward function. That is, given a joint state-action pair $(S, A)$ generated by a policy network as the input to the random distillation network, the reward is calculated by \eqref{eq2} described below.

\begin{equation}
\begin{split}
\begin{aligned}
&RD\_joint\_reward\left(S, A|\tau_{G}\right) =\\
&\exp \left(-v\left(f_{\hat{\theta}}\left(S, A|\tau_{G}\right)-f_{\theta}\left(S, A|\tau_{G}\right)\right)^{2}\right)
\label{eq2}
\end{aligned}
\end{split}
\end{equation}

\noindent where $v$ is a hyper-parameter, and $\tau_{G}$ is the generated trajectory of the joint state-action pair. The reason for defining such a reward function is that given an input pair $(S, A)$, if differences between output of the two networks is large, it indicates that the distribution the input pair follows is far from the distribution implied in the joint expert trajectories, therefore, the reward received should be low. In Lepus, $RD\_joint\_reward$ will be shared by multiple car agents. 

Then we note that the MADRaS platform provides a basic reward function (i.e., $g\_reward$ in Fig.~\ref{fig1}) for each car agent. This reward function is shown in \eqref{eq3}.

\begin{equation}
g\_reward = V_x cos(\alpha) - V_x sin(\alpha) - V_x |trackPos|
\label{eq3}
\end{equation}

\noindent where $V_x$ denotes the velocity along the longitudinal axis of the car agent, $\alpha$ denotes the angle between the direction of the car agent and the direction of the racing track axis, and $trackPos$ denotes the distance between the car agent and the track axis. The goal of this reward function is to maximize the axial velocity (i.e., $V_x cos(\alpha)$), minimize the lateral velocity (i.e., $V_x sin(\alpha)$), and penalize the car agent if it continues to deviate from the track center (i.e., $V_x |trackPos|$).

Lastly, as shown in \eqref{eq4}, we use the product of $RD\_join\_reward$ and $g\_reward$ as the final reward $R\_lepus_i$ to guarantee that each car agent $i$ can obtain a stable and reliable reward for every step it takes. This reward is then fed back into the policy networks to update their parameters.

\begin{equation}
\begin{split}
\begin{aligned}
& R\_lepus_i = g\_reward_i \cdot RD\_joint\_reward \\
& s.t. \ i = 1,2,...,M
\end{aligned}
\end{split}
\label{eq4}
\end{equation}

\subsection{Pre-training the Policy Network via an  Adversarial Process}
In order to speed up the training of policy networks, we adopt the adversarial networks structure for the pre-training of policy networks. Specifically, we first design a joint discriminator network as shown in Fig.~\ref{fig1}. Input data of this network consists of two parts, the joint state-action pairs extracted from the joint expert trajectory data, and another kind of joint state-action pairs, where the state is sampled from the joint expert trajectory data but the action is the output of the policy network using the sampled state as input. We train the joint discriminator network to distinguish these two parts of data.

In particular, the joint discriminator network consists of two neural networks with the same network structure and shared parameters, i.e., $Dis_\omega(S,A|\tau_E )$ and $Dis_\omega (S^{\prime},A_p|\tau_E )$, where $\tau_E$ denotes the joint state-action pairs obtained from the joint expert trajectories, $S^{\prime}$ denotes the set of states sampled from the joint expert trajectories, and $A_p$ denotes the set of actions generated from the policy networks based on $S^{\prime}$.

Either $Dis_\omega(S,A|\tau_E )$ or $Dis_\omega (S^{\prime},A_p|\tau_E )$ contains an input layer, two fully connected layers and an output layer. The numbers of hidden units of the two fully connected layers are $300$ and $600$, respectively, and $tanh$ is used as the activation function. The goal of the joint discriminator network is to maximize the function shown in \eqref{eq5}.

\begin{equation}
\hat{E}_{\tau_{E}}\left[\log \left(\operatorname{Dis}_{\omega}(S, A)\right)\right]+
\hat{E}_{\tau_{E}}\left[\log \left(1-\operatorname{Dis}_{\omega}(S^{\prime}, A_p)\right)\right]
\label{eq5}
\end{equation}

This equation is to maximize the gap between $Dis_w(S^{\prime}, A_p)$ and $Dis_w(S, A)$. However, the goal of pre-training the policy networks is to continuously reduce the gap between these two so that the output of two networks cannot be told apart. Therefore, the opposite of $Dis_{\omega}(S^{\prime}, A_p)$ is used as the loss function for pre-training the policy networks, as shown in \eqref{eq6}.

\begin{equation}
Loss_{policy}=-Dis_\omega(S^{\prime}, A_p)
\label{eq6}
\end{equation}

Based on \eqref{eq5} and \eqref{eq6}, we update the parameters of the discriminator network first and then policy networks, thus performing an adversarial training. Further, we iterate the above process many times to achieve the pre-training of policy networks, where the times (denoted by $Dis\_iter$) is a hyper-parameter.

\begin{figure*}[htbp]
\centering{\includegraphics[width=0.8\textwidth]{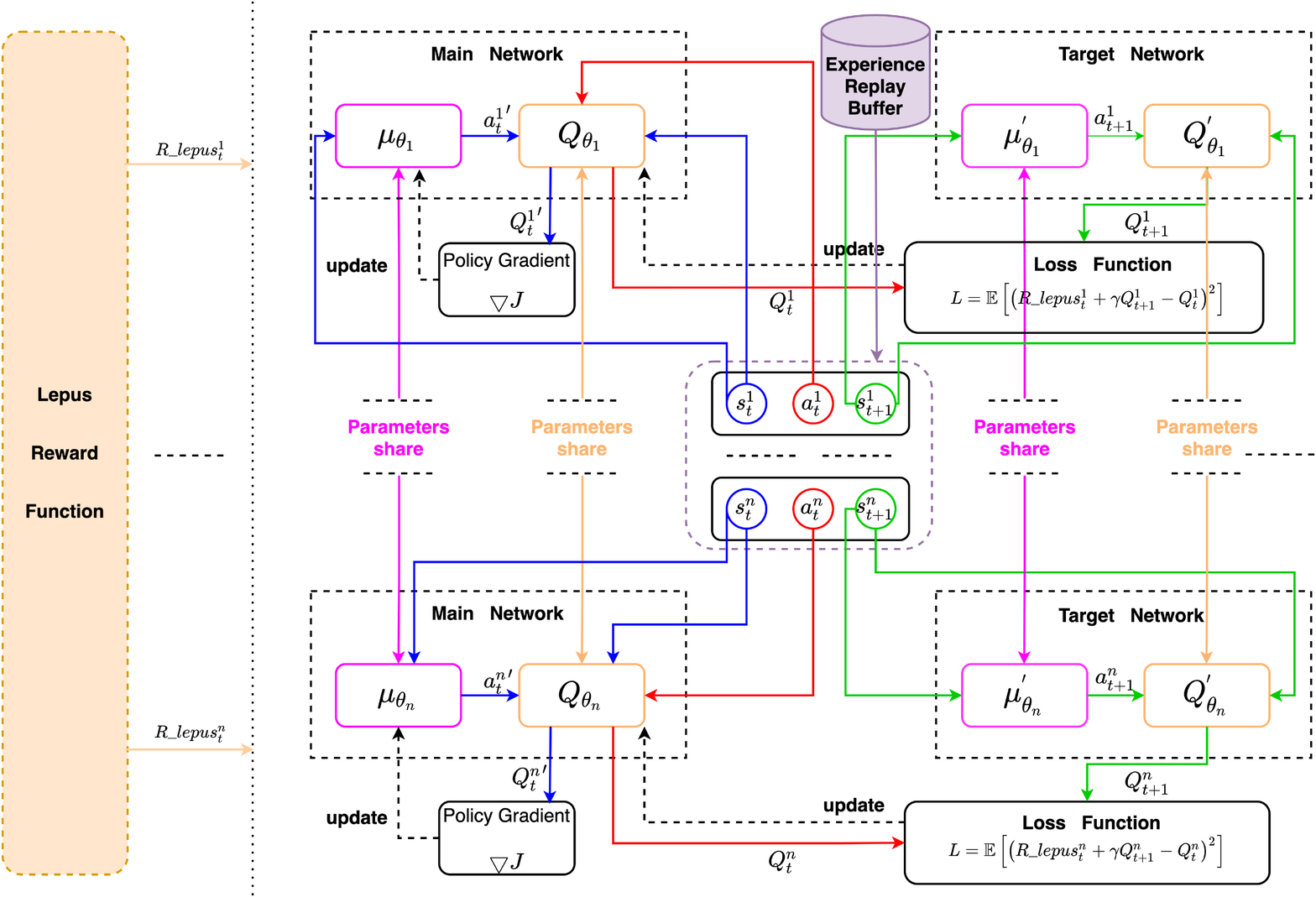}}
\caption{The training process of the policy networks with shared parameters.}
\label{fig2}
\end{figure*}

\subsection{Training Shared Parameters in Policy Networks}
All the policy networks share the same network parameters, each of which contains an input layer, two fully connected layers and an output layer. We set the numbers of hidden units of the two fully connected layers to $300$ and $400$, respectively, and choose ReLU as the activation function. The output of each policy network is a $3$-dimensional vector [steering, acceleration, braking] mentioned in Section $III$.

Additionally, the joint $Q$ network $\text{Joint}_{Q_{\theta}}$ and the policy networks share similar network structure and parameters, except that hidden units of the two fully connected layers of $Q$ network are $300$ and $600$, respectively.

The training method of the policy networks with shared parameters is inspired by the MADDPG algorithm. For each policy network, the input is a local states $s$, and the output is a deterministic action $a$ with O-U (Ornstein-Uhlenbeck Process) noise added. Then, all the local states $s$ and the actions $a$ generated from $s$ of all policy networks are stored in the experience buffer. After that, they are randomly sampled and feed into each $Q$ network to obtain the $Q$ value. The parameter updating of each policy network is achieved by maximizing the $Q$ value of each $Q$ network. When the training is done, the policy networks with shared network parameters make decisions for the car agents. The whole training process is shown in Fig.~\ref{fig2}.

The training process of a policy network is unstable if we only use a $Q$ network. To solve this problem, we create the target $Q$ network $Q_{\theta}^{\prime}$ and the target policy network $\mu_{\theta}^{\prime}$ by replicating the $Q$ network and the policy network. Then, we extract $\left(s_{t}^{j}, a_{t}^{j}, s_{t+1}^{j}\right)$ with a mini-batch of $N(j=\{1,...,N\})$ to obtain the ground truth of $Q$ (i.e., $y_{t}^{j}$), as shown in \eqref{eq7}, where $\gamma$ is the cumulative discount reward factor. And the loss function $L$ used in the parameter updating of the $Q$ network is shown in \eqref{eq8}.

\begin{equation}
\begin{split}
\begin{aligned}
& y_{t}^{j}={R\_lepus}_{t}^{j}+\gamma Q_{i}^{\mu^{\prime}}(s_{t+1}^{j},a_{t+1}^{j})|_{a_{i}^{j^{\prime}}=\mu_{i}^{j^{\prime}}(s^{j})} \\
& s.t. \ i = 1,2,...,M
\end{aligned}
\end{split}
\label{eq7}
\end{equation}

\begin{equation}
L=\frac{1}{N} \sum_{j}\left(y_{t}^{j}-Q_{i}^{\mu}\left(s_{t}^{j}, a_{t}^{j}\right)\right)^{2}
\label{eq8}
\end{equation}

The parameter updating method of the policy network is shown in \eqref{eq9}, while the parameter updating method of each target network is shown in \eqref{eq10}.

\begin{equation}
\begin{split}
\begin{aligned}
&\nabla_{\theta_{i}} J \approx \\
&\frac{1}{N} \sum_{j} \nabla_{\theta_{i}} 
\mu_{i}\left(s^{j}\right) \nabla_{a_{i}} 
Q_{i}^{\mu}\left(s^{j},a^{j}\right) 
|_{a_{i}=\mu_{i}\left(s^{j}\right)}
\label{eq9}
\end{aligned}
\end{split}
\end{equation}

\begin{equation}
\theta_{i}^{\prime} \leftarrow \ell \theta_{i}+(1-\ell) \theta_{i}^{\prime}
\label{eq10}
\end{equation}

\noindent where $\ell \in(0,1)$ is the parameter indicating the proportion of soft updating~\cite{lillicrap2015continuous}.

\section{Evaluation}
We implement Lepus with Tensorflow, adopting Adam for optimization. We evaluate Lepus on the MADRaS simulation platform. In our evaluation experiments, we train the Lepus using the hyper-parameters shown in Table \ref{table1}. The demo video is provided at https://github.com/qwerasdfzxcv852/demo-Lepus.

\begin{table*}[htbp]
\renewcommand{\arraystretch}{1.4}
\caption{Hyper-parameters.}
\begin{center}
\begin{tabular}{|c|c|c|c|c|}
\hline
\multicolumn{2}{|c|}{Network} & Hyper-parameter &  Description & Value\\
\hline
\multirow{6}*{Pre-training} & \multirow{4}*{Random distillation network} & RD\_iter & Training steps & $200$ \\
\cline{3-5}
\multirow{6}*{~} & \multirow{4}*{~} & RD\_minibatch & Mini-batch size & $256$ \\
\cline{3-5}
\multirow{6}*{~} & \multirow{4}*{~} & RD\_learning\_rate & Adam learning rate & $1e-3$ \\
\cline{3-5}
\multirow{6}*{~} & \multirow{4}*{~} & $v$ & Coefficient in RD$_{\text{reward}}$ & $250000$ \\
\cline{2-5}
\multirow{6}*{~} & \multirow{2}*{Discriminator network} & Dis\_iter & Training steps & \makecell[c]{$20000/3~ cars$\\$30000/4~cars$} \\
\cline{3-5}
\multirow{6}*{~} & \multirow{2}*{~} & Dis\_learning\_rate & Adam learning rate & $1e-4$ \\
\hline
\multirow{5}*{Joint training} & \multirow{2}*{Policy networks} & PN\_learning\_rate & Adam learning rate & $1e-4$ \\
\cline{3-5}
\multirow{5}*{~} & \multirow{2}*{~} & $\ell$ & Proportion of parameter soft updating & $1e-3$ \\
\cline{2-5}
\multirow{5}*{~} & \multirow{3}*{Q networks} & QN\_learning\_rate & Adam learning rate & $1e-3$ \\
\cline{3-5}
\multirow{5}*{~} & \multirow{3}*{~} & QN\_minibatch & Mini-batch size & $32$ \\
\cline{3-5}
\multirow{5}*{~} & \multirow{3}*{~} & $\gamma$ & Cumulating reward discount factor & $0.99$ \\
\hline
\end{tabular}
\label{table1}
\end{center}
\end{table*}

\subsection{Scenario Description}
MADRaS is a multi-agent extension of  TORCS~\cite{2013TORCS}. It is open source and mainly oriented toward multi-agent simulation training. The MADRaS simulation platform obeys the following basic driving rules:

\begin{itemize}
\item Each car agent should not run out of the racing track.
\item Each car agent should not accelerate along the opposite direction for a long time, which avoids driving backwards on the racing track.
\item Each car agent should not turn retrograde on the racing track.
\item During a specified time steps of the racing, the speed of the car agent should not be lower than a threshold or reduce to zero (We set one time step to $1$s, the number of time steps to $500$ and the speed threshold to $1$km/h).
\end{itemize}

If any of the rules above is violated, a training round of the cooperative driving of the multiple car agents ends. Besides, if all the cars run a lap around the racing track of the multi-agent cooperative driving, then a training round of multiple car agents is supposed to be ended.

We choose the CG Speedway No. 1 racing scenario under the quick race mode to conduct evaluation experiments, where one lap is 2057.56 meters.

\subsection{Joint Expert Trajectory Extraction}
Since Lepus works for the pure cooperative driving mode among multiple car agents, it is not necessary to obtain the corresponding expert trajectory data for each car agent. After treating all car agents as a whole, we just need to extract the joint expert trajectories for model training. The joint expert trajectory data have the characteristics that multiple car agents drive cooperatively in a given scenario, the distance between them is basically the same, and they cannot collide with each other during the entire scenario.

The steps of obtaining the joint expert trajectory data in the CG Speedway No. $1$ racing track scenario for our experiments are as follows. As shown in Fig.~\ref{fig1}, the joint expert trajectory data are extracted from the multi-agent PID (Proportional-Integral-Derivative) control model~\cite{willis1999proportional}. 

In general, PID control in the time domain can be expressed in \eqref{eq11}.

\begin{equation}
u(t)=K_{p} e(t)+K_{l} \int_{0}^{t} e\left(t^{\prime}\right) d t^{\prime}+K_{d} \frac{de(t)}{dt}
\label{eq11}
\end{equation}

\noindent where $e(t)$ denotes the deviation between desired value and actual value, and $K_{p}, K_{l}$ and $K_{d}$ denote the proportional, integral and derivative coefficients, respectively.

In Lepus, the goal of PID control is to guarantee that the generated actions of each car agent can make the car drive at a desired speed in a desired position of the track. The way of implementing PID control is to control the value of $V_x$ and $\alpha$ mentioned in \eqref{eq3}. Correspondingly, $e(t)$ in \eqref{eq11} will be the error function $V_x^{error}$ of car velocity or the angle error function $\alpha^{error}$. When the control begins, $V_x$ is initialized to $50km/h$ and $\alpha$ is initialized to 0. During the control, $V_x^{error}$  is calculated using the formula in \eqref{eq12}, and $\alpha^{error}$  is calculated using the formula in \eqref{eq13}.

\begin{equation}
{V_x}^{error} = 50.0 - V_x
\label{eq12}
\end{equation}
\begin{equation}
\alpha^{error} = - {trackPos}/{10} + \alpha
\label{eq13}
\end{equation}

Then, these two errors are feed into \eqref{eq11} to obtain the action vector [steering, acceleration, braking] of each car agent in each step of driving, where the acceleration value is calculated from ${V_x}^{error}$, the steering value is calculated from $\alpha^{error}$, and the braking value, initially set to 0, will be set to 1 when the acceleration value is less than 0.

In particular, we set the number of expert cars to three or four and extract $1000$ rounds of joint expert trajectory data from the CG Speedway No. $1$ racing track. We set the target driving speed to $50$km/h, therefore the time steps for the car to complete a round on the CG Speedway No. $1$ racing track are between $700$ and $800$. Then we filter the round data whose time step is less than $750$ from all the $1000$ rounds of data and obtain the final joint expert trajectory data for our experiments.

\subsection{Performance Comparison}
\subsubsection{Baselines and Experiments}
Based on the three different modes of multi-agent reinforcement learning methods, we choose the following reinforcement learning methods as competitors, including:

\begin{itemize}
\item DDPG-FDE, which adopts the fully decentralized mode. 
\item PS-DDPG, which adopts the fully centralized mode.
\item MADDPG, which adopts the centralized training and decentralized execution mode.
\end{itemize}

We   also   choose   an   imitation   learning   method   MAG-AIL(DDPG),  which  replaces  the  generator  in  the  originalMAGAIL with DDPG, as a competitor. The reason choosing MAGAIL(DDPG) instead of MAGAIL(PPO) is: (1) DDPG supports continuous actions and can satisfy the needs of the autonomous driving. (2) DDPG can solve the problem of the low sample utilization rate existed in GAIL-based methods, while PPO does not have this capability.

We conduct two groups of experiments where the car agents trained by Lepus and four baseline models are run and the number of cars is three or four, respectively.

\subsubsection{Performance Metrics}
For evaluating the performance of car agents, we define the stability of the cooperative driving of multiple car agents as the ratio of the driving distance to the number of collisions, as shown in \eqref{eq14}. That is, in given rounds, longer driving distance and fewer collisions stand for a higher stability.

\begin{equation}
Stability = \frac{travel\_distance}{1+ number\_ of\_collisions}
\label{eq14}
\end{equation}

\subsubsection{Result Analyses}

During the experiments, we set one lap around the racing track to a round, collect the number of collisions and the driving distance in a round for each car agent, and also collect the longest driving distance that the car agent runs in a round. Then we sum up experimental data of 20 rounds and calculate the average as well as the stability. Experimental results of three and four car agents are shown in Tables \ref{table2}-\ref{table3}, respectively. For every method, we sort and number the agents by their driving distances. 

From the results about driving distances, we find that Lepus outperforms other four methods in $20$ rounds of driving. 
Taking four car agents as an example, Lepus runs $693\%$, $358\%$, $397\%$, and $67\%$ longer than DDPG-FDE, MADDPG, MAGAIL(DDPG) and PS-DDPG on the average. We believe that the good performance of Lepus comes from the policy networks which have the shared parameters and are pre-trained.

\begin{table}[htbp]\scriptsize
\renewcommand{\arraystretch}{1.4}
\caption{Results of the first group of experiments: three car agents. The number in a bold type is the best performance in each row.}
\begin{center}
\begin{tabular}{|p{22pt}<{\centering}|p{16pt}<{\centering}|p{20pt}<{\centering}|c|p{26pt}<{\centering}|p{24pt}<{\centering}|p{23pt}<{\centering}|}
\hline
\multicolumn{2}{|c|}{Expt./Group 1} & \makecell[c]{DDPG\\-FED} & \makecell[c]{MAGAIL\\(DDPG)} & MADDPG & \makecell[c]{PS-\\DDPG} & Lepus\\
\hline
\multirow{2}*{Agent-I} & Distance & 4368.76 & 5635.40 & 6458.36 & 25739.80 & \textbf{41151.20}\\
\cline{2-7}
\multirow{2}*{~} & Collision & \textbf{0} & 220 & 920 & 120 & 1480 \\
\hline
\multirow{2}*{Agent-II} & Distance & 4190.62 & 5463.96 & 6389.32 & 24566.20 & \textbf{41151.20} \\
\cline{2-7}
\multirow{2}*{~} & Collision & \textbf{700} & 1380 & 1260 & 1880 & 860 \\
\hline
\multirow{2}*{Agent-III} & Distance & 4132.48 & 5411.38 & 6295.34 & 24476.40 & \textbf{39891.00} \\
\cline{2-7}
\multirow{2}*{~} & Collision & \textbf{700} & 1240 & 1220 & 1880 & 1560 \\
\hline
\multicolumn{2}{|c|}{Avg. Distance} & 211.50 & 275.20 & 319.10 & 1246.40 & \textbf{2036.56}\\
\hline
\multicolumn{2}{|c|}{Avg. Max. Distance} & 218.44 & 281.77 & 322.92 & 1286.99 & \textbf{2057.56}\\
\hline
\multicolumn{2}{|c|}{Avg. Collision No.} & \textbf{70.00} & 142.00 & 170.00 & 194.00 & 195.00 \\
\hline
\multicolumn{2}{|c|}{Stability} & 2.98 & 1.92 & 1.87 & 6.39 & \textbf{10.39}\\
\hline
\end{tabular}
\label{table2}
\end{center}
\end{table}

\begin{table}[htbp]\scriptsize
\renewcommand{\arraystretch}{1.4}
\caption{Results of the second group of experiments: four car agents. The number in a bold type is the best performance in each row.}
\begin{center}
\begin{tabular}{|p{22pt}<{\centering}|p{16pt}<{\centering}|p{20pt}<{\centering}|c|p{26pt}<{\centering}|p{24pt}<{\centering}|p{23pt}<{\centering}|}
\hline
\multicolumn{2}{|c|}{Expt./Group 2} & \makecell[c]{DDPG\\-FED} & \makecell[c]{MAGAIL\\(DDPG)} & MADDPG & \makecell[c]{PS-\\DDPG} & Lepus\\
\hline
\multirow{2}*{Agent-I} & Distance & 5360.08 & 8052.48 & 8441.32 & 22947.40 & \textbf{41151.20}\\
\cline{2-7}
\multirow{2}*{~} & Collision & 40 & 1182 & 938 & \textbf{0} & 20 \\
\hline
\multirow{2}*{Agent-II} & Distance & 5327.28 & 7571.52 & 8179.80 & 21905.00 & \textbf{38716.80} \\
\cline{2-7}
\multirow{2}*{~} & Collision & 20 & 1312 & 1504 & 2240 & \textbf{0}\\
\hline
\multirow{2}*{Agent-III} & Distance & 3856.10 & 7474.10 & 7588.30 & 21272.60 & \textbf{34600.00}\\
\cline{2-7}
\multirow{2}*{~} & Collision & 420 & 1148 & 1438 & 780 & \textbf{0} \\
\hline
\multirow{2}*{Agent-IV} & Distance & 3758.84 & 6121.52 & 7494.34 & 20640.40 & \textbf{30605.60}\\
\cline{2-7}
\multirow{2}*{~} & Collision & 420 & 966 & 430 & 2240 & \textbf{0} \\
\hline
\multicolumn{2}{|c|}{Avg. Distance} & 228.80 & 365.20 & 396.30 & 1084.60 & \textbf{1813.42} \\
\hline
\multicolumn{2}{|c|}{Avg. Max. Distance} & 268.00 & 402.62 & 422.07 & 1147.37 & \textbf{2057.56} \\
\hline
\multicolumn{2}{|c|}{Avg. Collision No.} & 45.00 & 230.40 & 215.50 & 263.00 & \textbf{1.00} \\
\hline
\multicolumn{2}{|c|}{Stability} & 4.97 & 1.58 & 1.83 & 4.11 & \textbf{906.71} \\
\hline
\end{tabular}
\label{table3}
\end{center}
\end{table}

From the results about collision number, we find that given three car agents, DDPG-FDE outperforms other four methods but its car agents often terminate half way because of triggering the driving rules. The fact illustrates that DDPG-FDE does not really outperform Lepus while taking both driving distance and collision number into consideration. Further, given four car agents, Lepus which adopts the 30000-step pre-training has the fewest collisions.

Lastly, as shown in the Stability line of Tables \ref{table2}-\ref{table3}, Lepus outperforms all other four methods in terms of the stability.

\subsection{Ablation Study}
We conduct the ablation study to observe the contributions of pre-training. Before that, we observe the effect of the following parts: the network structure of discriminator, the amount of expert trajectories used during the pre-training, as well as steps of pre-training.

First, we compare the impact of two different discriminator structures on performance. Discriminator A is the one finally adopted by Lepus. Discriminator B contains an input layer, three fully connected layers and an output layer. The numbers of hidden units of the three fully connected layers are $128$, $64$ and $32$, respectively, and $leakyReLU$ is used as the activation function. Columns 2-3 of Table \ref{table4} show the self-driving performance using different discriminators and all else being the same. That is, $50\%$ expert trajectories are used for pre-training and the pre-training steps are set to 20000.

Next, we adopt discriminator A and set the training steps to 20000 and observe the impact of different amount of the expert trajectories used in pre-training on the performance. The results are shown in columns 4-6 of Table \ref{table4}.

Then, we adopt discriminator A, and $50\%$ expert trajectories for pre-training and observe the impact of different pre-training steps on the performance. The results are shown in columns 7-9 of Table \ref{table4}.

\begin{table*}[htbp]
\renewcommand{\arraystretch}{1.3}
\caption{Performance under different experimental setups.}
\begin{center}
\begin{tabular}{|c|c|c|c|c|c|c|c|c|c|}
\hline
\multicolumn{2}{|c|}{\multirow{2}*{Expt./Experimental setup}} & \multicolumn{2}{c}{Discriminator} & \multicolumn{3}{|c|}{Amount of expert trajectories} & \multicolumn{3}{c|}{Pre-training step}\\
\cline{3-10}
\multicolumn{2}{|c|}{\multirow{2}*{~}} & A & B & $30\%$ & $50\%$ & $100\%$ & $10000$ & $20000$ & $50000$\\
\hline
\multirow{2}*{Agent-I} & Distance & 2057.56 & 13.10 & 609.54 & 2057.56 & 516.94 & 3.50 & 2057.56 & 816.42\\
\cline{2-10}
\multirow{2}*{~} & Collision & 74 & 31 & 17 & 74 & 67 & 0 & 74 & 40\\
\hline
\multirow{2}*{Agent-II} & Distance & 2057.56 & 20.06 & 600.64 & 2057.56 & 495.35 & 3.21 & 2057.56 & 811.65\\
\cline{2-10}
\multirow{2}*{~} & Collision & 43 & 38 & 31 & 43 & 67 & 0 & 43 & 40\\
\hline
\multirow{2}*{Agent-III} & Distance & 1994.55 & 29.86 & 474.46 & 1994.55 & 476.44 & 3.50 & 1994.55 & 462.43\\
\cline{2-10}
\multirow{2}*{~} & Collision & 78 & 28 & 20 & 78 & 29 & 0 & 78 & 0\\
\hline
\multicolumn{2}{|c|}{Stability} & 10.39 & 0.21 & 8.14 & 10.39 & 3.03 &  3.40 & 10.39 & 8.60\\
\hline
\end{tabular}
\label{table4}
\end{center}
\end{table*}

\begin{table*}[htbp]
\renewcommand{\arraystretch}{1.4}
\caption{Ablation study of Lepus.}
\begin{center}
\begin{tabular}{|c|c|c|c|c|c|c|c|}
\hline
Method & \multicolumn{3}{c|}{Driving distance of three cars} &  \multicolumn{3}{c|}{Collision number of three cars} & Stability\\
\hline
Lepus & 2057.56 & 2057.56 & 1994.55 & ~~~74~~~ & ~~~43~~~ & 78 & 10.39\\
\hline
Lepus-p & 1018.00 & 1038.02 & 1015.90 & 131 & 108 & 34 & 3.74\\
\hline
\end{tabular}
\label{table5}
\end{center}
\end{table*}

Lastly, given three car agents, we compare the performance of Lepus and Lepus-p, where Lepus equips with discriminator A, and uses $50\%$ expert trajectories for 20000-step pre-training, and Lepus-p denotes Lepus without pre-training. Table \ref{table5} shows the results. We can see that the performance of Lepus-p gets worse as expected, no matter whether the metric is distance, collision or stability.

\section{Conclusion}
In this paper, for the cooperative multi-car scenarios, we propose Lepus, a multi-agent cooperative decision-making method. We adopt the pure cooperative manner, design a random distillation network to learn an approximate reward function from the joint expert trajectories, and pre-train the policy networks via an adversarial process. In this way, Lepus can not only solve the sparse reward problem in realistic scenarios with incomplete information, but also improve the stability of car driving. Meanwhile, the parameter-sharing technique cuts down the training cost. Our experimental results show that Lepus outperforms all other four methods.

\section*{ACKNOWLEDGMENT}
This work was supported in part by the National Key Research and Development Project of China under Grant 2020YFB2103900.





\bibliographystyle{IEEEtran}
\bibliography{myref}





\end{document}